\documentclass{article}
\usepackage{ijcai21}

\RequirePackage[T1]{fontenc} 
\RequirePackage[tt=false, type1=true]{libertine} 
\RequirePackage[varqu]{zi4} 

\RequirePackage[libertine]{newtxmath}
\usepackage[letterspace=-25]{microtype}

\usepackage{soul}
\usepackage{url}
\usepackage[hidelinks]{hyperref}
\usepackage[utf8]{inputenc}
\usepackage[small]{caption}
\usepackage{graphicx}
\usepackage{amsmath}
\usepackage{booktabs}
\usepackage{savesym}
\savesymbol{Bbbk}
\savesymbol{openbox}

\usepackage{amsmath}
\usepackage{amsthm}
\usepackage{amsfonts}
\usepackage{hyperref}
\usepackage{float}
\usepackage{graphicx}
\usepackage{tikz}
\usepackage{stackengine}

\usepackage{amssymb}
\usepackage{amsthm}
\usepackage{MnSymbol}
\usepackage{color}
\usepackage{url}
\usepackage{multirow}
\usepackage{xcolor}
\usepackage{dsfont}
\usepackage{algorithm}
\usepackage[noend]{algpseudocode}

\restoresymbol{NEW}{Bbbk}
\restoresymbol{NEW}{openbox}

\usepackage[switch]{lineno}

\usepackage{tablefootnote}

\usepackage{subcaption}

\usepackage{todonotes}
\usepackage{tcolorbox}
\usepackage{amsfonts}
\usepackage{paralist}

\newcommand{\model}{\ensuremath{M}}

\newcommand{\inx}{\ensuremath{x}}
\newcommand{\ce}{\ensuremath{x'}}

\newcommand{\inputspace}{\mathcal{X}}
\newcommand{\outputspace}{\mathcal{Y}}
\newcommand{\R}{\mathbb{R}}

\DeclareMathOperator*{\argmin}{arg\,min}

\newcommand{\FT}[1]{\textcolor{black}{#1}}
\newcommand{\FL}[1]{\textcolor{black}{#1}}
\newcommand{\AR}[1]{\textcolor{black}{#1}}
\newcommand{\JJ}[1]{\textcolor{black}{#1}}

\usepackage{fancyhdr}
\fancypagestyle{firstpage}{%
    \fancyhf{}
    \setlength{\headsep}{0.3in}
    \cfoot{\thepage}
}
\fancypagestyle{otherpages}{%
    \fancyhf{}
    \setlength{\headsep}{0.3in}
    \rhead{Paper Title}
    \lhead{Authors}
    \cfoot{\thepage}
}

\title{Robust Counterfactual Explanations in Machine Learning: A Survey}

\author {
    Junqi Jiang,
    Francesco Leofante,
    Antonio Rago,
    Francesca Toni\\
\affiliations\large
    Department of Computing, Imperial College London, UK\\
\emails
    \{junqi.jiang, f.leofante, a.rago, f.toni\}@imperial.ac.uk
}

\begin{document}
\maketitle

\begin{abstract}
\emph{Counterfactual explanations (CEs)} are advocated as being ideally suited to providing algorithmic recourse for subjects affected by the predictions of machine learning models. While CEs can be beneficial to affected individuals, recent work has exposed severe issues related to the \emph{robustness} of state-of-the-art methods for obtaining CEs. Since a lack of robustness may compromise the validity of CEs, techniques to mitigate this risk are in order. In this survey, we review works in the rapidly growing area of \emph{robust CEs} and perform an in-depth analysis of the forms of robustness they consider. We also discuss existing solutions and their limitations, providing a solid foundation for future developments. 
\end{abstract}

\section{Introduction}
\label{sec:intro}

As the field of explainable AI (XAI) has matured, \emph{counterfactual explanations} (CEs) have emerged as one of the dominant post-hoc methods for explaining
AI models (see, e.g.~\cite{
KarimiBSV23} for an overview). CEs are often advocated as a means to provide \emph{recourse} for individuals that have been impacted by the decisions of a \FT{machine learning} model. In particular, given an input $\inx$ to a model $\model$, a CE essentially presents a user with a new, slightly modified input $\inx'$, which shows how a different outcome could be achieved if the proposed changes were to be applied to $\inx$. For illustration, consider a fictional loan application with features \emph{income} \textsterling$50K$, \emph{loan term} 35 months and \emph{loan amount} \textsterling$10K$ 
being rejected by 
\FT{a} model. \FT{In this example,} a CE 
could demonstrate 
that increasing \FT{the} \emph{income} to \textsterling$55K$  would result in the application being accepted. 

Given the critical nature of many scenarios in which CEs are 
\FT{deployed}, e.g. in financial or medical settings, it is of utmost importance that the recourse they provide is valid, i.e. it gives the intended change in outcome, and thus trustworthy. However, recent work has demonstrated that state-of-the-art 
methods \FT{for obtaining CEs} host major drawbacks when it comes to the \emph{robustness}, i.e. the validity under changing conditions, of the CEs they generate. In particular,~\cite{PawelczykAJUL22} showed that popular approaches for generating CEs may return explanations that are indistinguishable from adversarial examples. Broadly speaking, this means that CEs are extremely susceptible to small changes occurring in the setting they were generated for. To understand the implications of this, let us return to our loan example: if increasing income to \textsterling$55.1K$ (as opposed to the exact amount recommended) does not result in the application being accepted, the applicant may begin to question whether the CE was actually explaining the decision making of the AI model and was not just an artefact of the explainer instead. Worryingly, this is just one of many examples in which a lack of robustness in CEs may compromise their reliability.

In this survey, we conduct the first\footnote{To the best of our knowledge, only~\cite{JPMsurvey} have surveyed robustness in XAI methods previously. However, their primary focus was on feature attribution explanations instead of CEs.}comprehensive analysis of techniques developed to ensure that CEs are robust. After introducing the necessary background on CEs in Section~\ref{sec:counterfactuals}, we 
detail the methodology for our systematic survey of the existing literature on Robust CEs in Section~\ref{sec:survey}. We then classify approaches based on the form of robustness they consider and discuss each resulting category in Sections~\ref{sec:modelchanges}-\ref{sec:inputchanges}. In Section~\ref{sec:discussion}, we 
discuss key findings of the survey and look ahead to 
future \FT{prospects} in this emerging research field.

\section{Counterfactual Explanations}
\label{sec:counterfactuals}

Assume a classification model $\model:\inputspace \rightarrow 
\outputspace$ where $\inputspace$ is the input space and 
$\outputspace$ 
is a discrete set of output labels\footnote{Most approaches focus on binary classification ($\outputspace\!=\!\{0,1\}$).}. 
We focus on models $\model$ obtained by machine learning. Given an input point $\inx \in \inputspace$, most machine learning 
\FT{methods} (e.g. logistic regression, neural networks, tree ensembles
) first produce a \emph{class score} $s$ ranged in $[0, 1]$ for each class in $\outputspace$, 
and \FT{then} use it to determine the model classification (e.g. by choosing the class with the highest score). Given an input $\inx \in \inputspace$ and a model $\model$,
existing approaches compute a CE $\ce$ as \FT{follows}:
\begin{equation}
    \underset{\ce\in \inputspace}{\argmin} \text{ } cost(\inx, \ce) \text{ s.t. } \model(\ce) \neq \model(\inx)
\label{eqn:ceformulationexact}
\end{equation}
where $cost:\inputspace \times \inputspace \rightarrow \R^+$ is a suitable metric in the input space (e.g. $\ell_1$ or $\ell_{\infty}$ norm) \FT{and $\model(\ce) \neq \model(\inx)$ amounts to \emph{validity} of the CE}. In practice, solving the formulation exactly, e.g. through Mixed Integer Programming (MIP) as in~\cite{MohammadiKBV21}, may be viable only for certain types of classifiers. For differentiable classifiers, a relaxed formulation is typically considered instead:
\begin{equation}
    \underset{\ce\in \inputspace}{\argmin} \text{ } loss(\model(\ce), \model(\inx)) + \lambda \cdot cost(x, x')
\label{eqn:ceformulationgradient}
\end{equation}
where $loss$ is a differentiable loss function that pushes the search towards a valid CE $\ce$, i.e. one for which $\model(\ce) \neq \model(\ce)$, and $\lambda$ is a parameter dictating the trade-off term between validity and cost. Other properties have also been considered in the CE literature (see \cite{KarimiBSV23} for a recent survey of CEs and their properties). For instance, \emph{actionability} requires the CEs to act only on mutable features in realistic directions, \emph{causality} means that the changes in the input should conform with a full or a partial structural causal model, \emph{diversity} advocates 
the generation of a \emph{set} of different CEs for each input instead of a single CE, and \emph{plausibility} means a CE should be close to the data manifold.

\section{Systematic Survey}
\label{sec:survey}


We conducted a systematic search of the existing literature on robustness of CEs. We first performed keyword searches on Google Scholar
using the following patterns for exact matches: \emph{robust counterfactual explanation}, \emph{robustness of counterfactual explanation}, \emph{counterfactual explanation robustness}. For better coverage in each pattern, the words \emph{consistent} (\emph{consistency}) and \emph{stable} (\emph{stability}) were used in addition to \emph{robust} (\emph{robustness}). The phrase \emph{counterfactual explanation} was also interchangeable with \emph{counterfactual explanations, counterfactuals, recourse, algorithmic recourse}. To narrow down the topic, we filtered for technical papers from and after year 2017, when two seminal works in the CE literature were published \cite{DBLP:conf/kdd/TolomeiSHL17,wachter17}. For completeness, we also checked in Google Scholar for additional papers that cited the influential early works on robust 
\FT{CEs}, namely \cite{DBLP:conf/uai/PawelczykBK20,DBLP:conf/nips/UpadhyayJL21,DBLP:conf/nips/SlackHLS21}.

Table~\ref{table:bigtable} summarises the results of our survey. We found that the type of robustness falls within four separately studied categories, namely 
robustness against \emph{Model Changes} (MC), \emph{Model Multiplicity} (MM), \emph{Noisy Execution} (NE), and \emph{Input Changes} (IC). 
For the \FT{identified} approaches to these problems, we detail the suitable types of models to be explained, the level of access to the model which is required, the computational method used,
whether formal guarantees of robustness are given, 
and other \FT{considered} properties
.

In the next sections, we review the problem 
definitions, evaluation metrics for robustness, solutions and theoretical results for each 
\FT{category of robustness}.

\section{Robustness against Model Changes}
\label{sec:modelchanges}
\begin{table*}[t!]
    \begin{center}
    
    \small
    \begin{tabular}{llccccc}
    
    \cline{3-7}
     & 
     & 
     \textbf{Models} & 
     \textbf{Access} & 
     \textbf{Method} & 
     \textbf{Guarantee} & 
     \!\!\!\textbf{Properties}\!\!\! 
     \\
    \hline
    

    MC\!\!\!\!\!\! &
    
    \multicolumn{1}{l}{\cite{DBLP:journals/corr/mochaourab2021}} & 
    SVM & 
    Pred. & 
    Search & 
    - & 
    - 
    \\    

     &
    \multicolumn{1}{l}{\cite{DBLP:conf/nips/UpadhyayJL21}} & 
    LM & 
    Grad. & 
    GD+RO & 
    - & 
    Ac  
    \\

     &
    \multicolumn{1}{l}{\cite{DBLP:conf/iclr/BlackWF22}} & 
    \!\!NN\!\! & 
    Grad. & 
    GD & 
    - & 
   
    -  \\

     &
    \multicolumn{1}{l}{\cite{DBLP:conf/iclr/BuiNN22}} & 
    LM & 
    Grad.  & 
    GD  & 
    Prob.  & 
     
    Di  \\

     &
    \multicolumn{1}{l}{\cite{DBLP:conf/fuzzIEEE/SilvaB22}} & 
    $*$ & 
    Pred. & 
    Search & 
    - & 
     
    Pl  \\

     &
    \multicolumn{1}{l}{\cite{DBLP:conf/icml/DuttaLMTM22}} & 
    TM & 
    Pred. & 
    Search & 
    Prob. & 
    
    Pl  \\

     &
    \multicolumn{1}{l}{\cite{DBLP:journals/access/FerrarioL22}} & 
    $*$ & 
    White & 
    DA & 
    - & 
     
    -  \\

     &
    \multicolumn{1}{l}{\cite{DBLP:journals/corr/forel2022robust}} & 
    TM & 
    White & 
    MIP & 
    Prob. & 
     
    Ac/Pl  \\

     &
    \multicolumn{1}{l}{\cite{DBLP:conf/uai/NguyenBNYN22}} & 
    DM & 
    Grad. & 
    GD & 
    - & 
     
    Pl \\

     &
    \multicolumn{1}{l}{\cite{DBLP:journals/corr/bui2023coverage}} & 
    * & 
    Pred. & 
    GD, MIP & 
    - & 
     
    Ac/Pl  \\

     &
    \multicolumn{1}{l}{\cite{DBLP:conf/cikm/GuoJCSY23}} & 
    DM & 
    White & 
    GD+RO & 
    - & 
    
    Ac \\

     &
    \multicolumn{1}{l}{\cite{DBLP:conf/icml/HammanNMMD23}}  & 
    NN & 
    Pred. & 
    Search & 
    Prob. & 
    
    Pl \\

     &
    \multicolumn{1}{l}{\cite{DBLP:journals/corr/jiangacml23}} & 
    \!\!NN\!\! & 
    White & 
    MIP+RO & 
    Det. & 
     
    Ac/Pl  \\
     
     &
    \multicolumn{1}{l}{\cite{DBLP:conf/aaai/JiangL0T23}} & 
    \!\!NN\!\! & 
    White & 
    MIP & 
    Det. & 
    
    -  \\

     &
    \multicolumn{1}{l}{\cite{DBLP:conf/icml/KrishnaML23}} & 
    LM & 
    Grad. & 
    GD & 
    Det. & 
    
    -  \\

     &
    \multicolumn{1}{l}{\cite{DBLP:conf/iclr/NguyenBN23}} & 
    * & 
    Pred. & 
    MIP & 
    Det. / LM & 
     
    Ac  \\

     &
    \multicolumn{1}{l}{\cite{DBLP:conf/iclr/PawelczykLBK23}} & 
    DM & 
    Grad. & 
    GD & 
    - & 
     
    -  \\

     &
    \multicolumn{1}{l}{\cite{DBLP:journals/corr/wang2023tcol}} & 
    * & 
    Pred. & 
    Search & 
    - & 
     
    Ac/Di/Pl  \\
    \hline


    MM\!\!\!\!\!\! &
    
    \multicolumn{1}{l}{\cite{DBLP:conf/uai/PawelczykBK20}} & 
    $*$ & 
    Pred. & 
    Search & 
    - & 
    
    Pl  \\

     &
    \multicolumn{1}{l}{\cite{DBLP:conf/kr/LeofanteBR23}} & 
    \!\!NN\!\! & 
    White & 
    MIP & 
    Det. & 
     
    -  \\

     &
    \multicolumn{1}{l}{\cite{jiang2024recourse}} & 
    $*$ & 
    Pred. & 
    CA & 
    Det. & 
     
    -  \\
    
    \hline


    NE\!\!\!\!\!\! &
    \multicolumn{1}{l}{\cite{DBLP:conf/pkdd/HadaC21}} & 
    TM & 
    White & 
    MIP & 
    - & 
    -   \\

     &
    \multicolumn{1}{l}{\cite{DBLP:conf/icml/Dominguez-Olmedo22}} & 
    DM & 
    Grad. & 
    GD+RO & 
    LM & 
     
    Ac/Ca  \\

     &
    \multicolumn{1}{l}{\cite{DBLP:journals/corr/sharma2022}} & 
    * & 
    Pred. & 
    Search & 
    - & 
 
    Di/Pl  \\
    
     &
    \multicolumn{1}{l}{\cite{DBLP:conf/pkdd/GuyomardFGBT23}} & 
    DM & 
    Grad. & 
    GD & 
    Prob. & 
     
    -   \\

     &
    \multicolumn{1}{l}{\cite{DBLP:conf/eumas/LeofanteL23}} & 
    NN & 
    White & 
    FV & 
    Det. & 
     
    Ac  \\

     &
    \multicolumn{1}{l}{\cite{DBLP:journals/corr/maragno2023}} & 
    NN, TM & 
    White & 
    MIP+RO & 
    Det. & 
     
    Ac/Pl  \\

     &
    \multicolumn{1}{l}{\cite{DBLP:conf/iclr/PawelczykDHKL23}} & 
    DM & 
    Grad. & 
    GD & 
    Prob. & 
     
    Ac  \\

     &
    \multicolumn{1}{l}{\cite{DBLP:conf/aistats/RamanMS23}} & 
    DM & 
    Grad. & 
    Sampling & 
    Prob. & 
    
    \!\!\!Ac/Ca/Di/Pl\!\!\!  \\

    &
    \multicolumn{1}{l}{\cite{DBLP:journals/ai/VirgolinF23}} & 
    $*$ & 
    Pred. & 
    GA & 
    - & 
   
    Pl \\

    \hline


    IC\!\!\!\!\!\! &
    \multicolumn{1}{l}{\cite{DBLP:conf/ssci/ArteltVVHBSH21}} & 
    $*$ & 
    White & 
    MIP & 
    Prob. / LM & 
     
    Pl  \\

     &
    \multicolumn{1}{l}{\cite{DBLP:conf/nips/SlackHLS21}} & 
    DM & 
    Grad. & 
    GD & 
    - & 
     
    -  \\



     &
    \multicolumn{1}{l}{\cite{DBLP:conf/cikm/WangQLGM23}} & 
    $*$ & 
    Pred. & 
    SAT & 
    - & 
     
    Di  \\

     &
    \multicolumn{1}{l}{\cite{DBLP:journals/eswa/ZhangCWL23}} & 
    $*$ & 
    Pred. & 
    Search & 
    - & 
     
    Pl  \\

     &
    \multicolumn{1}{l}{\cite{DBLP:journals/corr/leofanteaaai24}} & 
    $*$ & 
    Pred. & 
    Search & 
    Det. & 
     
    Di  \\

    
    \hline
    \end{tabular}
    
    \end{center}
    \caption{Robust CE methods, partitioned based on the targeted robustness form. We include:
    \begin{inparaenum}[\it (i)]
    \protect\item the types of \underline{models} they target (support vector machines (SVM), linear models (LM), neural networks (NN), 
    differentiable models (DM),  
    tree-based models (TM) and 
    model agnostic methods ($*$));
    \protect\item the model \underline{access} required (\underline{white} box,
    \underline{grad}ients and 
    \underline{pred}ictions); 
    \protect\item the 
    \underline{method} used (gradient descent (GD),
    robust optimisation (RO),
    mixed-integer programming (MIP),
    computational argumentation (CA),
    formal verification (FV),
    genetic algorithms (GA), 
    satisfiability solving (SAT) and data augmentation (DA)); 
    \protect\item whether a robustness \underline{guarantee} is provided (\underline{Det}erministic, \underline{Prob}abilistic, for linear models only (LM) and 
    none ($-$));
    \protect\item other \underline{properties} 
    satisfied by the method, including \underline{Ac}tionability, \underline{Ca}usality, \underline{Di}versity, and \underline{Pl}ausibility. Note that all methods consider validity and proximity by default, thus they are omitted.
    \end{inparaenum}
    }
    \label{table:bigtable}
    \end{table*}

As shown in Table~\ref{table:bigtable}, the majority of approaches on robustness of CEs in the literature focus on robustness against MC. Model changes are typically defined as modifications in the parameters of a machine learning model that do not alter its architecture. These changes are typically assumed to 
result 
from retraining on inputs drawn from a slightly shifted data distribution~\cite{rawal2020algorithmic}. However, model changes resulting from data deletion queries have also been studied~\cite{DBLP:conf/iclr/PawelczykLBK23,DBLP:conf/icml/KrishnaML23}.

Different characterisations of the notion of change have been proposed. 
\cite{DBLP:conf/nips/UpadhyayJL21} first considered \emph{plausible model changes}, which are defined as updates to model parameters whose magnitude is bounded by a small constant. 
The same notion is also considered in~\cite{DBLP:conf/iclr/BlackWF22,DBLP:journals/corr/jiangacml23,DBLP:conf/aaai/JiangL0T23}.
The assumption on bounded updates is lifted in~\cite{DBLP:conf/icml/HammanNMMD23}, where \emph{naturally-occurring model changes} are studied in the context of neural networks. In a nutshell, this notion allows for unbounded changes as long as the updated model preserves a similar behaviour around the input being explained.

\FT{The property of robustness against MC can be summarised as follows.} 

\begin{tcolorbox}
\vspace{-0.2cm}
{\sc Robustness against MC}. 

Assume an input $\inx$ and a model $\model$. Let $\ce$ be a CE for $\inx$. Robustness against MC requires that whenever the model $M$ \textbf{changes} to $M'$, and this change is \textbf{sufficiently small}, then  $\model(\ce) = \model'(\ce)$. 
\vspace{-0.2cm}
\end{tcolorbox}

Despite different definitions of what constitutes a (small) change in this setting, e.g., some approaches require explicitly that $M(x)=M'(x)$), all existing works agree that a lack of such \emph{robustness against MC} could be a problem for both the user and the explanation providers. Consider the loan example: after being rejected, the applicant would expect that when applying again, achieving a \textsterling55K annual salary would result in the application being successful, as captured by the CE. However, a lack of robustness to MC may result in this CE being invalidated by small updates to the machine learning model (e.g. by 
\JJ{retraining with} new data). Ultimately this may lead to the application being rejected 
\FL{despite \AR{the applicant} having implemented the CE. On the contrary, a robust CE should preserve its validity in such a scenario.}

\subsection{Robustness Metrics}
\label{ssec:mcmetrics}

A commonly used metric to assess the robustness of CEs against MC is \emph{Validity after Retraining} (VaR). Roughly speaking, VaR measures the percentage of CEs that remain valid after the model for which they were generated 
is updated.
Common update strategies involve retraining from scratch using shifted datasets \cite{DBLP:conf/nips/UpadhyayJL21} 
or different portions of the original dataset \cite{DBLP:conf/icml/HammanNMMD23}, incremental retraining \cite{DBLP:conf/aaai/JiangL0T23}, and retraining with varying initialisation conditions \cite{DBLP:conf/iclr/BlackWF22}. This metric is also referred to as~\emph{recourse outcome instability}~\cite{DBLP:conf/iclr/PawelczykLBK23} and~\emph{recourse reliability}~\cite{DBLP:conf/eaamo/FonsecaBABS23}, the latter being defined for CEs in a multi-agent multi-step setting. A probabilistic formulation of this property is also given in~\cite{DBLP:conf/iclr/BuiNN22}, where the authors compute lower and upper bounds on the probability of a CE remaining valid after retraining of linear models.

To quantify robustness against plausible model changes, the \emph{$\Delta$-robustness} metric was proposed in~\cite{DBLP:conf/aaai/JiangL0T23}. Intuitively, a CE is said to be $\Delta$-robust if its validity is preserved across the whole family of models that can be obtained after applying bounded model changes. Given a CE $x'$, this can be can be captured by requiring that $\model(\inx') = \model'(\inx')$ for all $\model'$ such that $\lVert Param(\model') - Param(\model) \rVert_p \leq \delta \}$ for fixed $p,\delta$ and $Param$ being the vectorisation of a model's parameters.

\emph{Counterfactual Stability} (CS) is a robustness metric proposed to capture naturally-occurring model changes. A formulation of CS was introduced in~\cite{DBLP:conf/icml/DuttaLMTM22} for tree ensemble models to capture the intuition that \emph{i.} the class score for a robust CE should be high and that \emph{ii.} the class score of the CE's neighbouring input should also be high with a small variance.  In practice, for a given CE, the CS score evaluates the 
mean and standard deviation of the class scores of samples from a Gaussian distribution centred at the CE. 
The metric was later specialised in~\cite{DBLP:conf/icml/HammanNMMD23} to target naturally-occurring models changes in neural networks.

\subsection{Algorithms} 
\label{ssec:mcsolutions}


\paragraph{Robust optimisation approaches.} Targeting plausible model changes, \cite{DBLP:conf/nips/UpadhyayJL21} solve a min-max problem that generates the best CE under the largest admissible model change.   
At each step, an inner maximisation routine finds the weight perturbation vector that increases the prediction loss to the greatest extent. Then, an outer minimisation loop updates the CE which optimises the overall loss (Eq~(\ref{eqn:ceformulationgradient}) under the previously computed worst-case perturbation). 
Although this formulation natively supports linear models only, the authors show that it can also be applied to non-linear models by leveraging linear surrogate models obtained, e.g. using LIME \cite{DBLP:conf/kdd/Ribeiro0G16}\footnote{Note that in practice, all CE methods applicable to linear models can be also used for non-linear models by approximating them with linear surrogate models, but any theoretical properties could be lost.}.

Similarly to this approach, \cite{DBLP:journals/corr/jiangacml23} present a constrained optimisation formulation of the same problem in the context of feed-forward neural networks with piece-wise activations. Using a MIP encoding instead of gradient-based optimisation, the authors are able to provide stronger robustness guarantees than~\cite{DBLP:conf/nips/UpadhyayJL21}. However, this comes 
\FT{with higher computational costs.}

\paragraph{Increasing class scores.} Other approaches generate robust explanations by guiding the search towards CEs that result in high class scores. 
For example, \cite{DBLP:conf/icml/KrishnaML23} leverage leave-k-out analysis to approximate the solutions to the robust CEs problem by finding CEs with higher class scores on the original model.
\cite{DBLP:journals/corr/forel2022robust} increase the required class scores in the CE search and give probabilistic guarantees for ensembles of convex base learners. Similarly, \cite{DBLP:conf/fuzzIEEE/SilvaB22} target CEs with high class scores by exploring the input's neighbouring points using k-associated optimal graph analysis. Meanwhile, \cite{DBLP:journals/corr/mochaourab2021,DBLP:journals/corr/wang2023tcol} construct application-dependent CE prototypes and select 
those with higher class scores as robust CEs.

Complementing these results, \cite{DBLP:conf/iclr/BlackWF22} show that for complex non-linear models, higher class scores alone may not be sufficient to ensure robustness. Therefore, additional requirements need to be imposed on the search to promote robustness. For instance, \cite{DBLP:conf/iclr/BlackWF22} require that the CE\FT{s} be located in a region with a low Lipschitz constant and propose a method that leverages this result during CE search. Instead, \cite{DBLP:conf/icml/DuttaLMTM22,DBLP:conf/icml/HammanNMMD23} generate robust CEs by requiring high class scores not only for the CE, but also for its neighbouring points. A first search algorithm exploiting this idea is presented in~\cite{DBLP:conf/icml/DuttaLMTM22} in the context of decision trees; a gradient-based version is later developed in~\cite{DBLP:conf/icml/HammanNMMD23}. Both methods provide theoretical probabilistic guarantees for their CEs' VaR  
Finally, \cite{DBLP:conf/aaai/JiangL0T23} propose to strengthen the class score requirement with a robustness test based on the notion of $\Delta$-robustness. The authors propose an algorithm that iteratively generates CEs with higher class scores until the candidate solution is certified to be robust. For the certification step, a MIP encoding is used to explore the space of model changes exhaustively.

\paragraph{Probabilistic modelling.} Model changes can also be approximated using probabilistic modelling techniques as done in~\cite{DBLP:conf/uai/NguyenBNYN22}, where kernel density estimation techniques are used to model the data distributions, 
along with a Gaussian mixture ambiguity set 
to capture the 
model shifts. A min-max optimisation problem is then formulated to maximise the worst-case validity probability of 
CE\FT{s} and solved using gradient-based optimisation. A similar approach is taken in~\cite{DBLP:conf/iclr/NguyenBN23} where the space of possible model shifts is modelled using Gaussian mixtures over model parameters. Finally,~\cite{DBLP:conf/iclr/BuiNN22} obtain a lower bound on the probability of VaR by modelling the mean and covariance matrix of some nominal distribution assumed for the model parameters. Their lower bound can be relaxed to a differentiable form, and thus can be added as an additional term to the loss function of the gradient-based CE search. 

\paragraph{(Re-)Training for robustness.} Training methods have been proposed to robustify the CE generation process. Augmentation-based training is proposed in~\cite{DBLP:journals/access/FerrarioL22}, where CEs are used alongside training instances to train neural network classifiers. The authors show empirical results pointing to the fact that augmentation increases the chances of CEs remaining valid under successive shifts in the data-generating distribution. However, this technique may cause imbalances in the data, since the CEs are not naturally-collected data points from the same distribution as other training data. \cite{DBLP:conf/cikm/GuoJCSY23} formulate a tri-level optimisation problem which aims to simultaneously train an accurate neural network for prediction purposes, for which robust CEs c
\FT{an} be generated by passing any input into another jointly trained neural network. Finally, \cite{DBLP:journals/corr/bui2023coverage} sample data points (with their prediction results) centred at the original model's decision boundary near the input, and perturb the data covariance matrix of these samples. Then, linear surrogate models which are aware of the potential decision boundary shifts in the original model can be trained using the data shifts. They show that CEs generated on such linear surrogates using non-robust methods are robust.

\section{Robustness against Model Multiplicity}
\label{sec:modelmultiplicity}

MM (also called \emph{predictive multiplicity}) refers to the phenomenon that for a single machine learning task, multiple near-optimal models can be trained with similar test accuracies \cite{Breiman_01,DBLP:conf/icml/MarxCU20,Black_22}. MM 
can be dealt with 
by aggregating the models towards a single \emph{aggregated prediction} for each input \cite{black2022selective,Black_22}. 
However, recent results have shown that these models could have distinct behaviours for the same individuals in terms of predictions and their explanations \cite{DBLP:journals/natmi/Rudin19,DBLP:conf/icml/CostonRC21}. 
This has important implications for CEs, with one being that a CE for an aggregated prediction may not even be valid for all of the aggregated models, which would likely confound users' expectations, leading to the following 
\FT{high-level} definition:

\begin{tcolorbox}
\vspace{-0.2cm}
{\sc Robustness against MM}.

Assume an input $\inx$ and a set of models $\mathcal{M} 
$. 
Let $agg(\mathcal{M},\inx)$ be an \textbf{aggregated prediction} for $\inx$, and $\ce$ be a CE for $agg(\mathcal{M},\inx)$.
Robustness against MM requires that $\ce$ is \textbf{valid across some subset} $\mathcal{M}' \subseteq \mathcal{M}$, i.e. for any $\model_i \in \mathcal{M}'$
,  $\model_i(\ce) \neq agg(\mathcal{M},\inx)$.

\vspace{-0.2cm}
\end{tcolorbox}

In the CE literature, \cite{DBLP:conf/uai/PawelczykBK20} observe that traditional CEs generated on one original model have a high probability of being invalidated by other models resulting from MM. In this setting, it is assumed 
that \textit{the prediction result is fixed to one produced by one of the original models}, and the robustness target is to guard the CE validity against potential future changes in the classifier: a robust CE should ideally remain valid under the alternative models. This form of robustness is very similar to robustness against MC. The differences are that the MM problem does not assume a data distribution shift causing the model shifts, and there are no assumptions on the model types and architectures for the original model on which the 
\FT{CEs are} generated and for the alternative models. 


Recent advances have also raised concerns about \textit{when the prediction result for one input is pending} and to be decided by all the models under MM, instead of assuming a result from one original model \cite{black2022selective}. \cite{jiang2024recourse} investigate the problem of deciding the prediction result together with the CEs, with the robustness of the latter driving the former. That is, the solution should be a subset of models under MM and their (valid) CEs, from which the prediction results are decided.

\subsection{Robustness Metrics}
\label{ssec:mmmetrics}

The robustness evaluation for the fixed prediction scenario is 
very similar to the VaR for robustness against MC problem, using 
validity under alternative models
. These models could be arbitrary models trained on (different portions of) the same dataset with high accuracy. Thus, the same validity metric gives an intuitive 
evaluation of how robust the CEs are. 

For the pending prediction scenario, \cite{jiang2024recourse} propose desirable properties to characterise the optimal 
resulting set of models and CEs. 
\textit{Non-emptiness} requires that the resulting set be not empty, and ideally it should contain more than one model and CE such that the prediction result is collectively determined by multiple models with their explanations (\textit{non-triviality}).
\textit{Model agreement} states that the models included in the solution set should have the same prediction result for the input. If these results are the same as the prediction results by the majority of models in the MM model set, then the solution is said to satisfy \textit{majority vote}. Meanwhile, \textit{counterfactual validity} requires that every selected CE 
remain\FT{s} valid on every selected model. Finally, \textit{counterfactual coherence} enforces that if a model is selected as part of the solution set, then its corresponding CE should also be included, and vice versa. 

\subsection{Algorithms}
\label{ssec:mmsolutions}

\paragraph{Fixed prediction scenario.} \cite{DBLP:conf/uai/PawelczykBK20} provide important theoretical results demonstrating that the plausible CEs within the data manifold are more likely to stay valid under alternative models, compared with the closest CEs found by optimising Eqs (\ref{eqn:ceformulationexact}) or (\ref{eqn:ceformulationgradient})
, and that there is a plausibility-cost tradeoff, indicating that the more robust CEs will admit higher costs. 
CEs empirically show increased (but much lower than 100\%) validity under alternative models for an existing plausible CE method compared with non-plausible methods, along with increased costs. However, the method is not deployed to produce robust CEs. \cite{DBLP:conf/kr/LeofanteBR23} propose a solution to generating CEs that are guaranteed to be valid for a set of ReLU neural networks. Their method builds a product construction combining all specified models into a single neural network, on which 
\FT{CEs} can be generated by MIP. They also prove that finding a CE that is valid across a set of piece-wise linear models is NP-complete. 

\paragraph{Pending prediction scenario.} Leveraging methods in computational argumentation,  \cite{jiang2024recourse} present a novel argumentative ensembling method for finding a subset of 
the models and their CEs which satisfies the properties mentioned in Section \ref{ssec:mmmetrics} (with the exception of majority vote). In essence, this consists in grouping similarly-behaving models together to eliminate conflicts therebetween, then finding the maximal group of models and CEs as the final output. Alternatively, one could first decide the prediction result for the input 
using ensembling methods like majority voting or the method of \cite{black2022selective}, then 
select the group of models and CEs with this prediction as the final solution set.

\section{Robustness against Noisy Executions}
\label{sec:noisyexecution}
Traditional methods compute CEs under the assumption that the user receiving them will follow them to the letter. In practice, achieving the prescribed feature values exactly might not always be feasible, especially for the continuous features. Continuing from the example given in Section \ref{sec:intro}, a CE may suggest a salary increase of \textsterling$5342.04$ but achieving this exact change 
could be beyond the loan applicant's control. In some other cases, even if the prescribed change is achievable, the user may inadvertently introduce some noise in the implementation of the recourse. Broadly speaking, robustness to NE requires that a CE be invariant to these small changes\AR{:}  

\begin{tcolorbox}
\vspace{-0.2cm}
{\sc Robustness against NE}. 

Assume an input $\inx$ and a model $\model$. Let $\ce$ be a CE for $\inx$. Robustness against NE requires that whenever a \textbf{small perturbation} $\sigma$ is applied to $\ce$, 
validity is not affected, i.e.  $\model(\ce)=\model(\ce+\sigma)$.
\vspace{-0.2cm}
\end{tcolorbox}

Different characterisations of $\sigma$ have been proposed in the literature. For instance, \cite{DBLP:conf/iclr/PawelczykDHKL23,DBLP:conf/aistats/RamanMS23} assume noise could be sampled from some probability distribution \cite{DBLP:conf/iclr/PawelczykDHKL23,DBLP:conf/aistats/RamanMS23}. A more conservative formulation is used in ~\cite{DBLP:conf/icml/Dominguez-Olmedo22,DBLP:conf/eumas/LeofanteL23}, where the authors take an adversarial ML and/or verification angle, and define $\sigma$ as a $p$-norm ball around the CE. A specialised version of the latter is given in~\cite{DBLP:journals/ai/VirgolinF23}, 
in which a set of feature-specific perturbation vectors is obtained from domain knowledge.

\subsection{Robustness Metrics}
\label{ssec:nemetrics}

\cite{DBLP:conf/iclr/PawelczykDHKL23} formalise the notion of \textit{invalidation rate} (IR) of a CE as the expectation of the difference between the predicted label of the CE and those predicted for its noisy variants, i.e. $\mathbb{E}_{\sigma}[\model(x') - \model(x'+\sigma)]$. 
When used to empirically evaluate CEs, the expectation component can be approximated by sampling a certain number of noise vectors. 

Alternatively, 
inspired by the adversarial robustness literature, \cite{DBLP:conf/icml/Dominguez-Olmedo22} evaluate robustness by adding perturbation vectors (found by adversarial attack methods) to 
CE\FT{s} to invalidate 
\FT{them}, and measuring the minimum magnitude of the successful attack. Then, the 
CE\FT{s are} 
said to be robust to perturbation vectors of up to this magnitude. Similarly, \cite{DBLP:conf/eumas/LeofanteL23} employ formal verification tools to check the \textit{local robustness} of a CE. In particular, a verification query is formulated to check if a CE remains valid within an $\infty$-norm ball around it. These notions are more conservative than the IR and complement it by capturing worst-case scenarios. 

\subsection{Algorithms} 
\label{ssec:nesolutions}

\paragraph{Robust optimisation and class scores.} 
\cite{DBLP:conf/icml/Dominguez-Olmedo22} propose 
different solutions for both linear and non-linear models. For the linear case, the authors prove that increasing class scores is a sufficient condition for finding robust CEs. 
Similar remarks are also made in~\cite{DBLP:conf/pkdd/HadaC21,DBLP:journals/corr/sharma2022}. In the non-linear case, a robust optimisation approach is proposed. This is similar to \cite{DBLP:conf/nips/UpadhyayJL21} for the MC problem (Section \ref{ssec:mcsolutions}), except that the worst-case perturbation now applies to the CE itself instead of 
the model parameters. \cite{DBLP:journals/corr/maragno2023} provide a similar formulation using a MIP encoding, finding CEs with robustness guarantees for both neural networks and tree ensemble models.

\paragraph{Verification-based approach.} 
\cite{DBLP:conf/eumas/LeofanteL23} show that local robustness queries as commonly phrased in adversarial machine learning can be used to evaluate a given CE's robustness. Then, the authors provide an iterative algorithm to quantify robustness by embedding a local robustness check into a binary search procedure.

\paragraph{Novel loss functions.}  \cite{DBLP:conf/iclr/PawelczykDHKL23} propose a method that allows the end user to specify the IR of their CEs. The intuition behind this is that robust CEs typically have higher cost, which the user may not be willing to sustain. Specifying a threshold on the IR allows the end user to control the level of risk they intend to take, thus implicitly reducing the cost from the most robust CEs. The authors give a differentiable first-order approximation of the IR for linear models and Gaussian noise applied to the CE. This term can then be added to the loss function of 
\FT{Eq}~(\ref{eqn:ceformulationgradient}) and optimised with gradient descent. Under specific assumptions on the choices of models and CE generation algorithms (e.g. logistic regression models and using the method of~\cite{wachter17} as the base CE method), IR can be expressed analytically, and a probabilistic robustness guarantee in terms of IR can be obtained through setting the class score. To tackle these strong assumptions and their inherent limitations on practicality, \cite{DBLP:conf/pkdd/GuyomardFGBT23} relax the definition of IR to capture class scores 
with which they provide a tighter upper bound on the exact IR. Then, the upper bound approximation by Monte Carlo estimation is plugged into the loss function and CEs are optimised in the same manner as in \cite{DBLP:conf/iclr/PawelczykDHKL23}. 
In \cite{DBLP:journals/ai/VirgolinF23}, a robustness term which is similar to IR is added to Eq~(\ref{eqn:ceformulationgradient}), and then minimised 
via a genetic algorithm.

\paragraph{Probabilistic approaches.} The methods of \cite{DBLP:conf/icml/Dominguez-Olmedo22,DBLP:journals/corr/maragno2023,DBLP:conf/eumas/LeofanteL23} effectively obtain a region containing valid CE points. Such regions could also be characterised by probability distributions, from which CEs could be sampled. \cite{DBLP:conf/aistats/RamanMS23} take a Bayesian hierarchical modelling approach where noise is explicitly modelled by random variables with Gaussian or scaled Dirichlet distributions, for continuous and categorical variables respectively. Their method outputs a posterior distribution which estimates the counterfactual distribution, and a Hamiltonian Monte Carlo sampling method is applied to generate diverse CEs. 

\section{Robustness against Input Changes}
\label{sec:inputchanges}

Unlike the previous notions of robustness 
which mostly focus on the validity of CEs under certain perturbations, robustness against IC requires that CEs generated for similar inputs are consistent. A first argument in favour of this property is given by \cite{DBLP:conf/fat/Hancox-Li20}, who 
advocates that explanations generated for similar inputs should not differ radically to improve the justifiability of explanations. However, \cite{DBLP:conf/nips/SlackHLS21} found that traditional CE generation algorithms may fail to satisfy this property, raising fairness concerns. As an example, the paper shows that neural networks can be trained in such a way that individuals belonging to a non-protected group can always obtain a lower cost recourse when compared to protected group. Following this example, we can formulate robustness against IC as follows.

\begin{tcolorbox}
\vspace{-0.2cm}
{\sc Robustness against IC}.

Assume two inputs $\inx_1, \inx_2$ and a model $\model$ such that $\model(\inx_1) = \model(\inx_2)$. Let $\ce_1,\ce_2$ be CEs for $\inx_1,\inx_2$, respectively. Robustness against IC requires that whenever $\inx_1, \inx_2$ are \textbf{similar}, then $\inx'_1,\inx'_2$ are also \textbf{similar}.

\vspace{-0.2cm}
\end{tcolorbox}

Similarity between inputs is typically defined in terms $p$-norm balls, e.g. given an input $\inx_1$, an input $\inx_2$ is similar to $\inx_1$ if $\lVert \inx_{1} - \inx_2 \rVert_p \leq \delta$. Then, one could use this formulation to derive a similarity notion for CEs, bounding the maximum distance therebetween. For example,~\cite{DBLP:journals/corr/leofanteaaai24} discuss the case where $\lVert \ce_1 - \ce_{2} \rVert_p \leq k \text{ } \lVert x_1 - x_2\rVert_p, k \in \R^+$. A special case of robustness against IC is encountered when dealing with inputs with missing feature values, where $\inx_2$ differs from $\inx_1$ only in the missing attributes. The returned CEs should ideally capture validity for any missing feature values, as discussed in~\cite{DBLP:journals/corr/kanamori2023}.

\subsection{Robustness Metrics}
\label{ssec:icmetrics}

Robustness measures are usually characterised by the expected distance between the CEs of similar inputs, $\mathbb{E}_{\inx_{2} \sim S}[d(\ce_1, \ce_{2})]$, where $S$ denotes a set of inputs that are similar to $x_1$ and $d$ is a distance metric defined over the input space. This quantity is identified as the \textit{local instability} of CEs \cite{DBLP:conf/ssci/ArteltVVHBSH21}, and is targeted by most of the surveyed studies. If the method generates a diverse set of CEs for a single input, then local instability needs to be generalised to account for distances between two sets of points, as 
in \cite{DBLP:conf/cikm/WangQLGM23,DBLP:journals/corr/leofanteaaai24}.

\subsection{Algorithms} 
\label{ssec:icsolutions}

\paragraph{Adversarial manipulations and defenses.} \cite{DBLP:conf/nips/SlackHLS21} 
propose an adversarial training framework whereby the cost of 
\FT{CE} changes can be made to change drastically for across protected and unprotected subgroup of data points. They demonstrate that such instabilities against small perturbations in the input are evident for traditional gradient-based CE methods, and discuss possible mitigation strategies. 
These include randomising the CE search initialisation, reducing the number of features used for CE search, or reducing the model size to limit overfitting.

\paragraph{Plausibility and robustness.} \cite{DBLP:conf/ssci/ArteltVVHBSH21} give some theoretical results about the upper bound of the local instability of CEs for linear binary classifiers under Gaussian and Uniform noise in the input. They show that plausible CEs, i.e. CEs that lie within the data manifold, exhibit a higher degree of invariance to input perturbations. This insight is also the intuition behind the method by \cite{DBLP:journals/eswa/ZhangCWL23} which aims to maximise robustness to input changes by finding more plausible CEs. Similarly, \cite{DBLP:conf/cikm/WangQLGM23} 
proposes a boolean satisfiability approach to generate plausible CEs, which also demonstrate a lower local instability than other traditional CE methods. 

\paragraph{Robustness via diversity.} The above methods do not explicitly target the local instability of CEs. \cite{DBLP:journals/corr/leofanteaaai24} show that satisfying robustness notions such as $\lVert \ce_1 - \ce_{2} \rVert_p \leq k \text{ } \lVert x_1 - x_2\rVert_p, k \in \R^+$ may be impossible for traditional CE methods in the general case. Therefore, the authors propose to move away from single-instance CEs and instead consider (diverse) sets of CEs. Under some assumptions, the authors prove that a relaxed form of robustness can be satisfied whereby sets of explanations generated for similar inputs are guaranteed to contain similar CEs.

\section{Summary and Outlook}
\label{sec:discussion}

We conducted a comprehensive and fine-grained analysis of
robust CE generation approaches and categorised them into the type of robustness they consider. 
In doing so, we identified some 
open research questions that are 
shared across the robustness spectrum. We discuss them in this section, providing an outlook for the future of this emerging research field.

\paragraph{Robustness vs cost trade-offs.} 
Several works discuss the existence of a trade-off between the cost of a CE and its robustness. Specifically, increasing cost is often discussed as a necessary condition for improving a CE's robustness~\cite{DBLP:conf/nips/UpadhyayJL21,DBLP:conf/iclr/PawelczykDHKL23}. Increasing the cost of a CE typically implies steering the CE search away from the decision boundaries of a model. This often results in increased class scores, which is a sufficient condition for increasing (some forms of) robustness in linear models~\cite{DBLP:conf/icml/Dominguez-Olmedo22,DBLP:conf/nips/UpadhyayJL21,DBLP:conf/iclr/BuiNN22}. However, the understanding of this trade-off within the context of non-linear models, such as neural networks, is still limited. For example,~\cite{DBLP:journals/corr/jiangacml23} 
show
that several robust methods often find less costly CEs than non-robust alternatives. 
Theoretical results have been obtained on bounding the maximum cost increase needed to achieve robustness~\cite{DBLP:conf/iclr/PawelczykDHKL23,DBLP:conf/pkdd/GuyomardFGBT23}. However, we argue that more research is needed to advance our understanding of robustness-cost trade-offs.

\paragraph{One form of robustness to fit them all?} Existing robustness notions may share some similarities, depending on the types of models and methods considered. For instance, when considering 
tree-based models, the problem of guaranteeing robustness against MC and MM tends to converge and algorithms developed for one setting might be applicable to the other. 
However, the same considerations may not apply to other robustness notions. For example, \cite{DBLP:journals/corr/maragno2023} target robustness against MC and NE and demonstrate that, at least in the case of linear models, these two notions are orthogonal in general. Furthermore, \cite{krishna2022on} highlights an interesting interplay between robustness of CEs and general adversarial robustness of a model, showing that popular training techniques to robustify neural networks may also lead to increased robustness in CEs. This begs the question as to whether existing robustness notions, including adversarial robustness, are strictly disjoint. However, 
their potential connections are largely unexplored. We advocate for more in-depth studies to shed light on this promising 
research area.

\paragraph{Links to fairness.} Some notions of CE robustness appear to have strong connections with existing literature on CE fairness. This is especially true for robustness against IC. In simple words, robustness against IC requires that \textit{similar CEs be given to similar individuals}, whereas CE fairness requires that \textit{recourse of similar cost be generated for similar individuals} (modulo differences on protected features). Therefore, robustness against IC only requires similarity between CEs without constraining their cost wrt specific features. Despite this difference, recent work by~\cite{DBLP:conf/fat/EhyaeiKSM23} showed that robustness often implies fairness, complementing existing results~\cite{DBLP:conf/nips/SlackHLS21,DBLP:journals/corr/gupta,DBLP:conf/aies/SharmaHG20,DBLP:conf/aaai/KugelgenKBVWS22} and paving the way for new research directions at the intersection between CEs, robustness and fairness.

\paragraph{Lack of standardised benchmarks.} 
Most works only include a limited number ($\leq2$) of robust CE baseline methods in their empirical studies. This is understandable since the topic emerged only recently, but it hinders understanding of how practical each method is. For example, optimisation methods based on MIP are able to generate exact solutions with strong guarantees, but this typically comes with higher computational cost. On the other hand, gradient-based methods may provide suboptimal results but can benefit from the highly-parallelised deep learning libraries. Additionally, several studies target more than one property 
concurrently, e.g. robustness and plausiblity~\cite{DBLP:conf/uai/PawelczykBK20,DBLP:journals/corr/jiangacml23}, which further complicates empirical comparisons. Going forward, we argue that intensive research effort should be put into developing standardised libraries to evaluate robustness. Similar initiatives have been pursued within the CE arena, e.g.~\cite{DBLP:conf/nips/PawelczykBHRK21}, and could provide a solid starting point for this.

\paragraph{Lack of user studies.} Robustness is typically framed as a functional requirement to be evaluated using mechanistic metrics. However, a lack of robustness in CEs has the potential to weaken their justifiability and thus jeopardise their explanatory function~\cite{DBLP:conf/fat/Hancox-Li20}. None of the studies reported in this survey have conducted user experiments to explore this phenomenon, which we argue should be given more prominence and potentially guide the development of novel algorithmic solutions in the future.

\section*{Acknowledgements}
{Jiang, Rago and Toni were partially funded by J.P. Morgan and by the Royal Academy of Engineering under the Research Chairs and Senior Research Fellowships scheme. 
Leofante is supported by an Imperial College Research Fellowship grant. 
Rago and Toni were partially funded by the European Research Council (ERC) under the European Union’s Horizon 2020 research and innovation programme (grant agreement No. 101020934). 
Any views or opinions expressed herein are solely those of the authors listed.}

\bibliographystyle{named}
\bibliography{bib}

\end{document}